%% file: main.tex
\documentclass{article} % For LaTeX2e
\usepackage{iclr2023_conference_tinypaper,times}
\iclrfinalcopy

\input{math_commands.tex}

\usepackage{amsmath,amssymb,amsfonts}
\usepackage{bm}
\usepackage{algorithmic}
\usepackage{graphicx}
\usepackage{textcomp}
\usepackage{xcolor}
\usepackage{tabularx}
\usepackage{multirow}
\usepackage[pagebackref=false, breaklinks=true, colorlinks, citecolor=citecolor, linkcolor=linkcolor, bookmarks=false]{hyperref}
\definecolor{citecolor}{HTML}{000000}
\definecolor{linkcolor}{HTML}{ED1C24}

\usepackage[capitalise,noabbrev]{cleveref}

\newcolumntype{Y}{>{\centering\arraybackslash}X}
\def\BibTeX{{\rm B\kern-.05em{\sc i\kern-.025em b}\kern-.08em
    T\kern-.1667em\lower.7ex\hbox{E}\kern-.125emX}}

\def\ie{\textit{i.e.}}
\def\eg{\textit{e.g.}}

\def\textemb{$\bm{e}_\text{text}$}

\renewcommand{\cite}{\citep}

\title{SD-NAE: Generating Natural Adversarial \\Examples with Stable Diffusion}

\author{Yueqian Lin\thanks{Equal contribution.}\\
    Duke Kunshan University \\
    \texttt{yueqian.lin@duke.edu} \\
    \And
    Jingyang Zhang\footnotemark[1], Yiran Chen, Hai Li \\
    Duke University \\
    \texttt{\{jz288, yiran.chen, hai.li\}@duke.edu}
}

\begin{document}

\maketitle

\vspace{-2mm}
\begin{abstract}
Natural Adversarial Examples (NAEs), images arising naturally from the environment and capable of deceiving classifiers, are instrumental in robustly evaluating and identifying vulnerabilities in trained models.
In this work, unlike prior works that \textit{passively} collect NAEs from real images, we propose to \textit{actively} synthesize NAEs using the state-of-the-art Stable Diffusion.
Specifically, our method formulates a controlled optimization process, where we perturb the token embedding that corresponds to a specified class to generate NAEs.
This generation process is guided by the gradient of loss from the target classifier, ensuring that the created image closely mimics the ground-truth class yet fools the classifier.
Named SD-NAE (Stable Diffusion for Natural Adversarial Examples), our innovative method is effective in producing valid and useful NAEs, which is demonstrated through a meticulously designed experiment.
%Robustly evaluating deep learning image classifiers is challenging due to some limitations of standard datasets.
%Natural Adversarial Examples (NAEs), arising naturally from the environment and capable of deceiving classifiers, are instrumental in identifying vulnerabilities in trained models.
%Existing works collect such NAEs by filtering from a huge set of real images, a process that is passive and lacks control.
%Our work thereby provides a valuable method for obtaining challenging evaluation data, which in turn can potentially advance the development of more robust deep learning models.
Code is available at \url{https://github.com/linyueqian/SD-NAE}.
\end{abstract}

\input{sections/1_introduction}

\input{sections/3_methodology}

\input{sections/4_results}
\input{sections/5_conclusion}


% \subsubsection*{Acknowledgements}
% Use unnumbered third level headings for the acknowledgements. All
% acknowledgements, including those to funding agencies, go at the end of the paper.
\newpage
\subsubsection*{URM Statement}
The authors acknowledge that at least one key author of this work meets the URM criteria of the ICLR 2024 Tiny Papers Track.
\subsubsection*{Acknowledgements}
The research was supported in part by NSF 1822085, the NSF IUCRC for ASIC and the memberships contributed by our industrial partners (\url{https://asic.pratt.duke.edu/}).

\bibliography{ref}
\bibliographystyle{iclr2023_conference_tinypaper}

\appendix
\input{sections/Appendix}

\end{document}

%% file: math_commands.tex
\usepackage{amsmath,amsfonts,bm}

\def\eqref#1{equation~\ref{#1}}
\def\1{\bm{1}}

\DeclareMathAlphabet{\mathsfit}{\encodingdefault}{\sfdefault}{m}{sl}
\SetMathAlphabet{\mathsfit}{bold}{\encodingdefault}{\sfdefault}{bx}{n}

%% file: sections/1_introduction.tex
\section{Introduction}

Robustly evaluating deep image classifiers is challenging, as existing standard test sets such as ImageNet \cite{deng2009imagenet} often feature simpler image compositions \cite{imagenetv2} and ``spurious features'' \cite{geirhos2018imagenettrained}, which can lead to an overestimate of model performance.
To address this issue, \citet{nae} introduce ``Natural Adversarial Examples'' (NAEs), where they employ adversarial filtration over extensive real images to pinpoint challenging natural images that deceive classifiers.
NAEs are valuable in assessing worst-case performance and uncovering model limitations.
Their approach to obtaining NAEs, however, is limited by its passive nature and lack of control over the selection of specific types of challenging examples, thereby restricting the ability to fully explore classifier weaknesses in diverse scenarios.

In this work, we propose to \textit{actively} synthesize NAEs with a controlled optimization process.
Leveraging a class-conditional generative model, particularly Stable Diffusion \cite{rombach2021highresolution},
we optimize the class token embedding in the condition embedding space.
This process is guided by the gradients of classification loss from the target image classifier to ensure the adversarial nature of the generated examples.
Our method, termed SD-NAE (Stable Diffusion for Natural Adversarial Example), not only achieves a non-trivial fooling rate against an ImageNet classifier but also offers greater flexibility and control compared to previous methods, highlighting SD-NAE's potential as a tool for evaluating and enhancing model robustness.
%and \textit{controlled} approach for generating NAEs using 
%Our method, termed SD-NAE (Stable Diffusion for Natural Adversarial Example), diverges from passive selection of NAEs and focuses instead on synthesizing NAEs through optimization of the class token embedding in the condition embedding space.
%This process, guided by the gradients of classification loss from the target image classifier, ensures the adversarial nature of the generated examples.
%Unique to our approach is the emphasis on perturbing the condition rather than the image's latent space, as in previous works \cite{uae}, and prioritizing the \textit{token embedding} optimization over \textit{text embedding}.
%This methodology not only achieves a non-trivial fooling rate against an ImageNet classifier but also offers greater flexibility and control compared to previous methods like \cite{nae}, highlighting SD-NAE's potential as a tool for evaluating and enhancing model robustness.

%% file: sections/3_methodology.tex
\section{Methodology}

\begin{figure*}[t]
\centering
\includegraphics[width=0.85\linewidth]{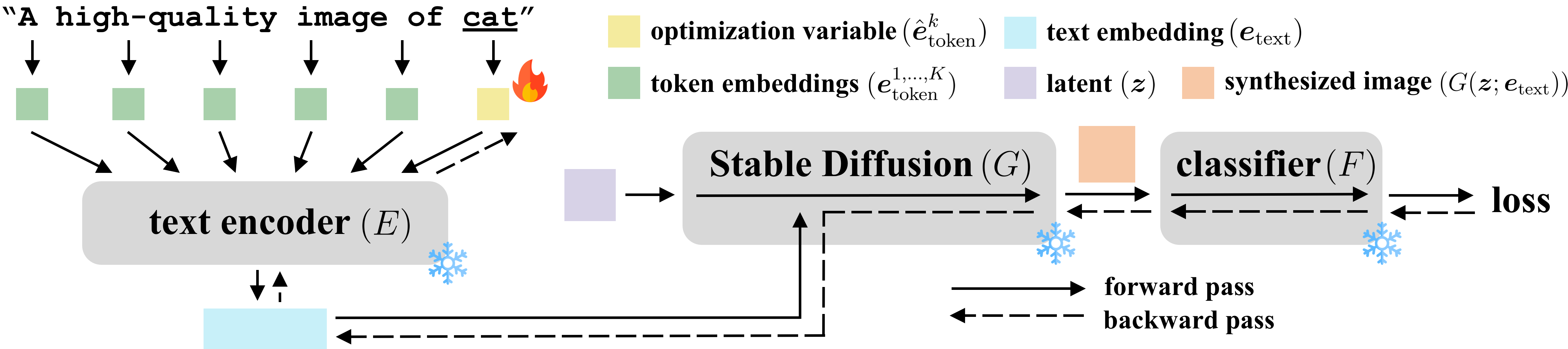}
   \vspace{-3mm}
   \caption{Guided by the loss gradient backpropagated from the classifier, SD-NAE generates NAEs by optimizing only the class-related token embedding, while keeping all models frozen. The letters in the parentheses are notations used in \cref{eq:sd-nae}.}
   %\vspace{-2mm}    
\label{fig:method}
\end{figure*}

We introduce the SD-NAE method (\cref{fig:method}), which is motivated by the concept of NAEs and uses Stable Diffusion to approximate natural-looking images (see \cref{sec:2_related_work} for background introduction).
Our exploration focuses on how adversarial optimization can enhance this approach, comparable to the creation of pixel-perturbed adversarial examples \cite{szegedy2013intriguing}.
The core of SD-NAE lies in the strategic optimization of class-relevant token embedding to trick the classifier into misclassifying the generated image. 
Consider, for instance, an image of a cat generated by Stable Diffusion $G$ following the text condition ``A high-quality image of a cat''.
Initially, the image can be correctly identified as a cat by the classifier $F$, given Stable Diffusion's accurate generation capability. 
However, through subtle alterations of the token embedding of ``cat'', we expect to induce misclassification (\eg, towards a target class $y$, $y\neq$``cat'') over the generated image while maintaining its ground-truth as ``cat''.
This process is governed by the optimization equation:
\begin{equation}
\min_{\hat{\bm{e}}_\text{token}^{k}}L(F(G(\bm{z};\bm{e}_\text{text})),y)+\lambda\cdot R(\hat{\bm{e}}_\text{token}^{k},\bm{e}_\text{token}^{k}),\text{where}~\bm{e}_\text{text}=E(\bm{e}_\text{token}^{0},...,\hat{\bm{e}}_\text{token}^{k},...,\bm{e}_\text{token}^{K-1}).
\label{eq:sd-nae}
\end{equation}
We mark the notations in \cref{fig:method}, while leaving a detailed discussion of \cref{eq:sd-nae} in \cref{sec:method_expand}.
In general, the optimization variable $\hat{\bm{e}}_\text{token}^{k}$ is the class-relevant token embedding (corresponding to ``cat'' in our example).
The first term encourages the produced image to be adversarial, while the second term makes sure that the perturbation on $\hat{\bm{e}}_\text{token}^{k}$ is only moderate, retaining the natural appearance of the generated image and preserving its ground-truth label as ``cat''.

%This delicate balance demonstrates the novelty of the method in generating adversarial examples while preserving the integrity of the original image class. 
%SD-NAE represents a novel approach in the field of adversarial optimization and offers potential for further research in image generation and classification.

%% file: sections/4_results.tex
\section{Experiment}

\begin{figure*}[t]
\centering
\includegraphics[width=0.85\linewidth]{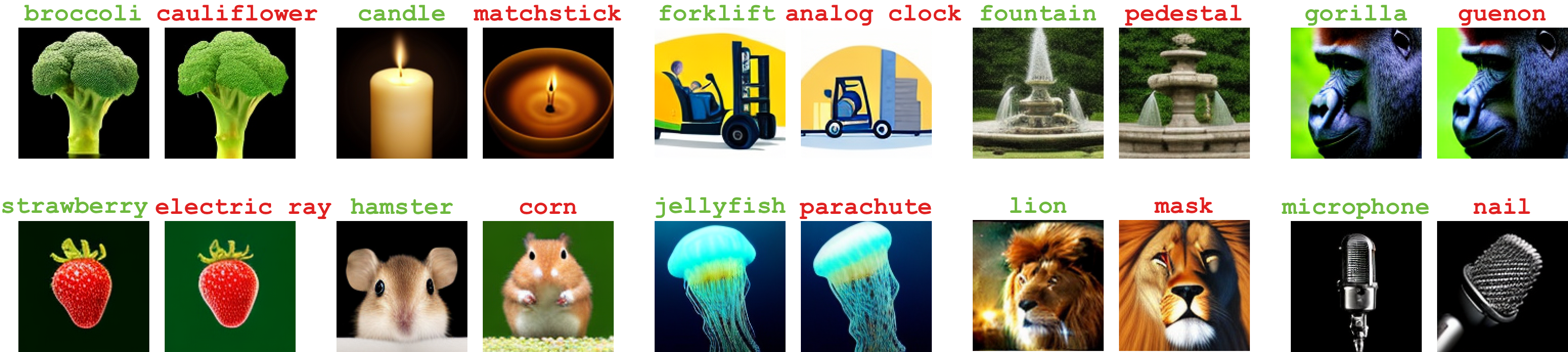}
   \vspace{-3mm}
   \caption{NAEs generated by SD-NAE. In each pair, the left initialization image is correctly classified by the model, yet the right one optimized by our method gets misclassified with the wrong prediction marked in red. See more samples in \cref{fig:results}.}
   %\vspace{-4mm}    
\label{fig:main_selected}
\end{figure*}

We evaluate SD-NAE using a carefully designed experiment.
Please see details in \cref{sec:expr_expand}.
Essentially, we focus on 10 categories of ImageNet whose semantics is clear.
We take them as the ground-truth and generate 20 samples using SD-NAE for each of the 10 classes, resulting in 20 * 10 = 200 total optimization processes.
An optimization is deemed successful if the image at any optimization step gets misclassified by the target classifier, which is a ResNet-50 pretrained on ImageNet.
Importantly, we make sure that all initialization images (prior to optimization by SD-NAE) are correctly classified.
In such a setting, our SD-NAE achieves a noteworthy success rate of 43.5\% (which is actually a lower bound; see \cref{sec:expr_expand}), demonstrating its capability to effectively generate NAEs.
Furthermore, as can be seen in \cref{fig:main_selected}, the images generated by SD-NAE display variations in color, background, view angle, and style, underscoring its potential as a tool for examining model generalization among various covariate shifts.
For comparison with the prior work of \citet{uae}, please see \cref{sec:compare}.

%Initially working with 100 random classes, we refined this number to 25 based on classifier accuracy, and further to 10 to facilitate clarity in human evaluation.
%We verified that the initially generated images were correctly classified by ResNet-50, thereby establishing a foundation for effective SD-NAE optimization.
%The efficacy of SD-NAE was quantitatively assessed using the `fooling rate' metric.
%This process entailed 200 optimization procedures across the selected classes, defining success as the ability to fool the classifier into misclassifying an image, yet maintaining its original classification to a human observer.
%SD-NAE achieved a fooling rate of 43.5\%, a noteworthy outcome given the classifier's initial perfect accuracy with these images.
%This demonstrates SD-NAE's capability to generate Natural Adversarial Examples effectively. 
%Furthermore, the images generated by SD-NAE displayed variations in color, background, and style, underscoring its potential as a tool for examining model generalization and image diversity.

%% file: sections/5_conclusion.tex
\section{Conclusion}

% In this work, we propose SD-NAE, a novel method that leverages Stable Diffusion to actively generate Natural Adversarial Examples.
% SD-NAE has a formulated optimization process that provides control and is flexible for adaptation.
% Quantitative and qualitative results demonstrate that SD-NAE effectively produces useful and interesting NAEs.

% Since SD-NAE is based upon Stable Diffusion, it inherits a few limitations from SD.
% First, the computational cost of SD-NAE could be high and the optimization could be slow.
% For instance, generating a single 128x128 image with SD-NAE requires approximately 22GB of GPU memory and takes about 1 minute; both the memory footprint and time cost escalate with increased image size and sampling steps.
% Second, in some cases, we find that the generated image is absurd and far from natural.
% However, we expect all these issues to be significantly alleviated as Stable Diffusion improves over time.

% From a broader perspective, using generative models to synthesize data for evaluation is a promising and valuable direction that extends beyond images to text and speech domains.
% As modern deep learning models grow in size and are trained on increasing amounts of data, they become increasingly powerful. This raises questions about the adequacy of existing public test datasets in revealing the weaknesses of these trained models.
% We hope that SD-NAE can motivate further studies in this direction.

SD-NAE, by leveraging Stable Diffusion, effectively generates Natural Adversarial Examples and demonstrates its significant potential in the field of robustness research.
As deep learning models continue to evolve, we believe that SD-NAE presents a novel approach for evaluating and understanding these complex systems, thereby emphasizing its profound role in future research.
We discuss related works and limitations of SD-NAE in \cref{sec:2_related_work} and \cref{sec:limitation}, respectively.

%Although it encounters challenges such as high computational costs and occasional unnatural outputs, we believe SD-NAE's contributions to data synthesis across diverse domains, including image, text, and speech, are substantial.

%% file: sections/Appendix.tex
\section*{\centering Appendix}

\section{Background and Related Work}
\label{sec:2_related_work}

In this section, we provide background information and discuss related works to facilitate the presentation of our method.
We first introduce the definition of NAEs, then distinguish our study from prior works on generating NAEs, and finally give an overview of the functionality of Stable Diffusion.

\subsection{Natural Adversarial Examples}
\label{sec:nae_concept}

Formally, NAEs are defined as a set of samples w.r.t. a target classifier $F$ \cite{nae,uae}:
\begin{equation}
    \mathcal{A}\triangleq\{\bm{x}\in\mathcal{S}|O(\bm{x})\neq F(\bm{x})\}.
\label{eq:nae}
\end{equation}
Here, $\mathcal{S}$ contains all images that naturally arise in the real world and look realistic to humans.
$O$ is an oracle that yields the ground-truth semantic category of the image and relies on human evaluation.

Notice that NAEs are less restricted than pixel-perturbed adversarial examples (often referred to as ``adversarial examples'') \cite{szegedy2013intriguing}, which are samples \textit{artificially} crafted by adding minor perturbations to image pixels.
Both NAEs and adversarial examples (AEs) can expose the vulnerability of a given classifier.
However, since AEs are artificial rather than natural, they are mostly studied in the security context where there is assumed to be an attacker that intentionally attempts to compromise a model.
In contrast, studies of NAEs, including ours, focus on a broader setting where the samples naturally occur within the environment but are misclassified by the classifier.

\subsection{Generating NAEs}
\label{sec:related_work}
As previously mentioned, we contend that the passive filtering of NAEs from real images, as demonstrated by \citet{nae}, lacks flexibility; in contrast, we utilize a generative model to synthesize NAEs.
In this regard, our work is closely related to but also has essential distinctions with the work by \citet{uae}.
While they optimize the latent of a class-conditional GAN with a fixed condition, we propose perturbing the condition while keeping the latent fixed within Stable Diffusion.
In fact, it is later found that GAN can be sensitive to the latents, and generated images may be of low quality when the optimized latents land outside the well-defined support \cite{dai2023advdiff}.
We will show that applying their concept to Stable Diffusion is less effective in producing NAEs compared to our method.

Building upon this comparison, it is pertinent to discuss a concurrent work by \citet{dai2023advdiff}, who also apply the diffusion model to generate NAEs.
However, their method is to enforce classifier guidance \cite{dhariwal2021diffusion} to be adversarial, which requires sophisticated modification to the default classifier-free guidance sampling \cite{ho2021classifierfree} and may need extra care to adapt to different samplers.
In contrast, our method can readily generalize to various diffusion models as it only perturbs the condition embedding without interfering with the sampling process.

\subsection{Stable Diffusion}
Stable Diffusion represents a family of latent diffusion models \cite{rombach2021highresolution} with the capability of conditional generation.
Using $G$ to represent the Stable Diffusion model, the formulation that best describes its functionality in the context of our work is
\begin{equation}
    \bm{x}=G(\bm{z};\bm{e}_{\text{text}}),
\end{equation}
where $\bm{x}$ is the synthesized image, $\bm{z}$ is a (random) latent vector, and $\bm{e}_\text{text}$ is the text embedding which serves as the condition for the generation.
More specifically, \textemb{} is the output of a transformer-based text encoder $E$ where the input is the token embedding sequence $\bm{e}_\text{token}^{0:K-1}$ corresponding to the raw text description after tokenization, \ie,
\begin{equation}
    \bm{e}_{\text{text}}=E(\bm{e}_\text{token}^0,\bm{e}_\text{token}^1,...,\bm{e}_\text{token}^{K-1}).
\end{equation}
where $K$ denotes the maximum padded length specified by the encoder $E$.

\begin{figure*}[th]
\centering
\includegraphics[width=0.55\linewidth]{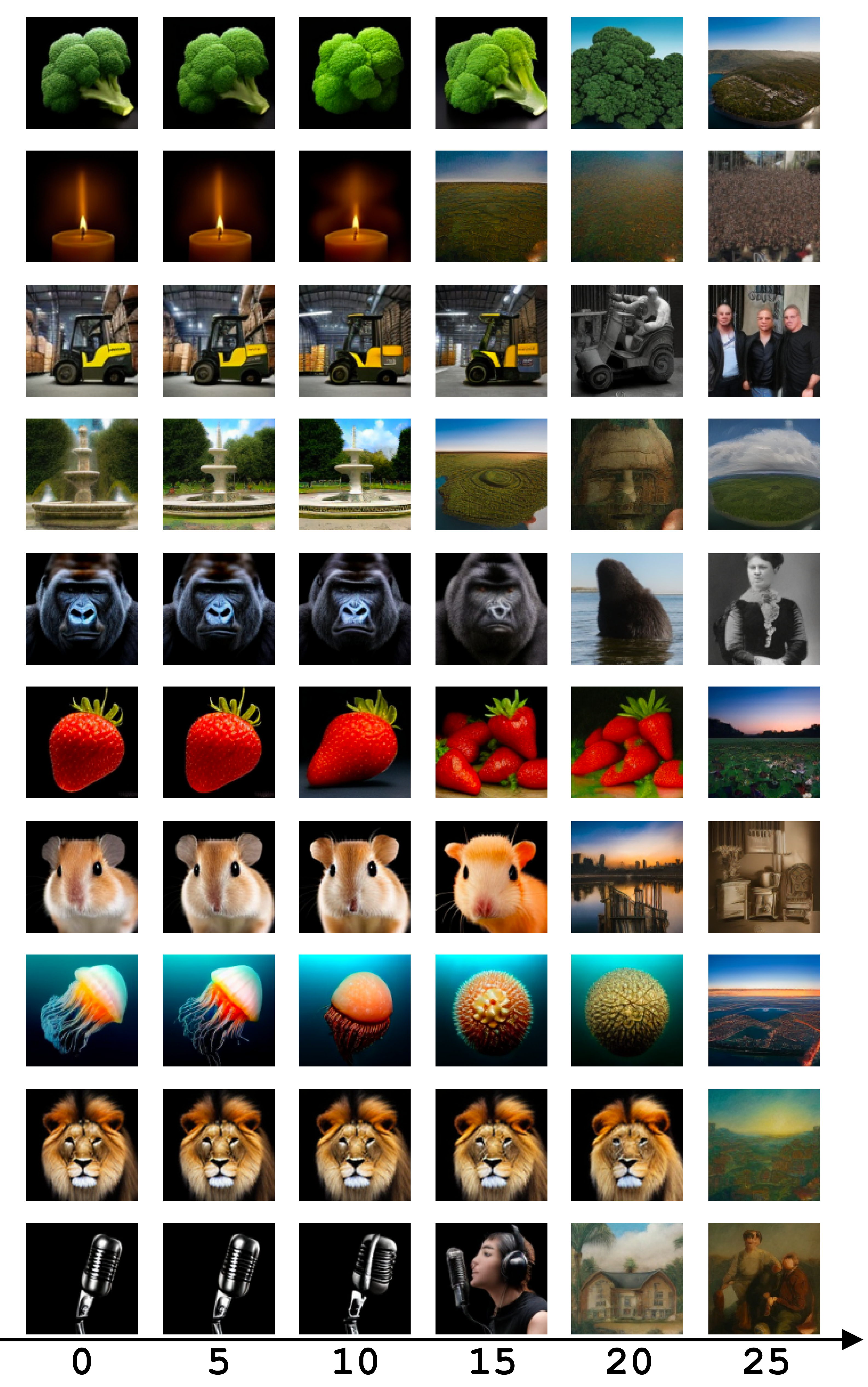}
   %\vspace{-1mm}
   \caption{Empirical evidence that constraining the magnitude of token embedding perturbation can help preserve the image ground-truth. From left to right of each row, we move the initialized class token embedding along a random yet fixed Rademacher vector (\ie, each element has equal probability of being +1 or --1) with increasing magnitude. The bottom axis denotes the relative magnitude of the perturbation, and the real magnitude has a factor of 1e-3. It can be noticed that the image semantic is well-preserved when the perturbation is small.}
   %\vspace{-4mm}
\label{fig:tk_embed_vicinity}
\end{figure*}

\section{Expanded Discussion on SD-NAE}
\label{sec:method_expand}

First, we provide a more detailed explanation of our optimization objective in \cref{eq:sd-nae}.
Let us explain the variables with a concrete example for clarity.
Suppose we want to generate an image of a cat with a text condition being ``A high-quality image of a cat'' (other prompts can also work here as long as it contains the keyword ``cat''), such that the image is misclassified by $F$ as some category other than ``cat''.
Equivalently and more formally, the goal is $O(G(\bm{z};\bm{e}_\text{text}))=\text{``cat''}\neq F(G(\bm{z};\bm{e}_\text{text}))$.

The optimization variable, denoted as $\hat{\bm{e}}_\text{token}^k$ in \cref{eq:sd-nae}, corresponds to the token embedding of the word ``cat''
We initialize $\hat{\bm{e}}_\text{token}^k$ with the original token embedding $\bm{e}_\text{token}^k$ and optimize it with two terms.
The first term aims to encourage the generated sample $G(\bm{z};\bm{e}_\text{text})$ to deceive the classifier $F$ into making an incorrect prediction.
Specifically, if we want to induce a targeted misclassification towards a specific class $y$ ($y\neq \text{``cat''}$), we can simply use the cross-entropy loss as $L$.
For untargeted misclassification, one can set $y$ to ``cat'' and use negative cross-entropy as $L$ (\ie, maximizing the classification loss of class ``cat'').

The second term serves to regularize $\hat{\bm{e}}_\text{token}^k$, ensuring that the ground truth $O(G(\bm{z};\bm{e}_\text{text}))$ remains unchanged during optimization; otherwise, the unintended equality $O(G(\bm{z};\bm{e}_\text{text}))=F(G(\bm{z};\bm{e}_\text{text}))$ may occur, contradicting the definition of NAEs.
To achieve this, we can let the regularization $R$ be a distance metric (\eg, Euclidean distance or cosine similarity) to enforce $\hat{\bm{e}}_\text{token}^k$ to stay in the vicinity of its unmodified counterpart $\bm{e}_\text{token}^k$.
In other words, we are only inducing moderate perturbation to the token embedding, which intuitively helps $O(G(\bm{z};\bm{e}_\text{text}))$ remain unchanged (\eg, being ``cat'' all the time in our example).
We empirically justify this intuition and design in \cref{fig:tk_embed_vicinity}.
The $\lambda$ that accompanies the second term is just a weighting factor.
It is worth noting, however, that in practice, when the number of optimization steps is small, it often suffices to set $\lambda=0$ since the overall magnitude of the perturbation (\ie, the distance between $\hat{\bm{e}}_\text{token}^k$ and $\bm{e}_\text{token}^k$) is bounded.

\textbf{Why token embedding?}
We next discuss why we choose the token embedding $\bm{e}_\text{token}^k$ instead of the latent $z$ or the text embedding \textemb{} as the optimization variable here.
Empirically, we find that perturbing the latent or text embedding is significantly less effective and efficient in generating NAEs compared to perturbing the token embedding, which has fooling rates of 10\%, 20\%, and 43.5\% respectively (refer to \cref{sec:expr_expand} for the definition of fooling rate and experiment setup).
Our hypothesis for this observation is as follows.
Firstly, in a diffusion model, the latent undergoes an iterative multi-step reverse diffusion process, potentially impeding the gradient flow from the classifier back to the latent. 
Secondly, as the text embedding integrates all tokens, perturbations on the text embedding might disperse across all tokens.
Intuitively, perturbing some class-irrelevant tokens (\eg, the meaningless padding tokens) is not likely to induce significant change to the image content, meaning that there is less chance the generated sample will fool the classifier.
In contrast, perturbing the class-relevant token (\ie, $\bm{e}_\text{token}^k$) directly targets the semantic-related content of the image, which we will further demonstrate to be effective with the following experiment results.

\textbf{Other application scenarios.}
Lastly, notice that SD-NAE can be easily adapted to create other types of NAEs beyond in-distribution (ID) misclassification.
For instance, a deployed model in the real world will inevitably encounter Out-of-Distribution (OOD) samples, which are samples not belonging to any known category \cite{Zhang_2023_WACV,zhang2023openood}, necessitating an accurate OOD detector to flag unknown inputs. 
With SD-NAE, one can generate NAEs that fool the OOD detector into ID~\textrightarrow~OOD or OOD~\textrightarrow~ID misclassification by playing with the loss $L$ in \cref{eq:sd-nae}.
Specifically, notice that an OOD detector often operates by thresholding the maximum softmax probability.
To generate an OOD image predicted as ID by the detector, we can employ cross-entropy loss as $L$ and use any one-hot label as $y$, thereby encouraging the classifier to make a confident prediction on the synthesized image.
Reversely, the ID~\textrightarrow~OOD misclassification is also achievable if we minimize the maximum classification probability by minimizing the cross-entropy between the softmax probability distribution and uniform distribution \cite{Zhang_2023_WACV}.
Overall, SD-NAE demonstrates flexibility in producing NAEs for various purposes.

\section{Experiment Details}
\label{sec:expr_expand}

\textbf{Models and setup.}
The target classifier, which we aim for the NAEs to fool is an ImageNet-pretrained ResNet-50 hosted by Microsoft on Hugging Face (model tag: \textsf{``microsoft/resnet-50''}).
We utilize a nano version of Stable Diffusion, finetuned from the official 2.1 release (model tag: \textsf{``bguisard/stable-diffusion-nano-2-1''}).
We generate 128x128 images to ensure the optimization is manageable with a single 24GB GPU; these images are then resized to 224x224 before being fed to the classifier, matching its default resolution.
DDIM sampler with 20 sampling steps is used for Stable Diffusion.
The guidance scale is set to the default value of 7.5.
In the optimization of SD-NAE, we use Adam as the optimizer with a learning rate of 0.001.
The number of iterations or gradient steps is 20.

\textbf{Workflow and metric.}
We use \textit{fooling rate} as the quantitative metric for SD-NAE.
To ensure a fair and meaningful evaluation, we first do several careful preprocessing as follows.
We start with 100 random classes from ImageNet.
For each class, we generate 100 samples from Stable Diffusion with random latent and measure the accuracy of the classifier on those samples.
Subsequently, we remove the classes whose accuracy is lower than 90\%, which leaves us 25 classes.
After that, we manually pick ten classes whose semantics are clear and unambiguous to our human evaluators (the authors of this work) to make it easy for later human inspection (to get the oracle prediction $O(\bm{x})$).
The selected classes are \texttt{broccoli}, \texttt{candle}, \texttt{forklift}, \texttt{fountain}, \texttt{gorilla}, \texttt{strawberry}, \texttt{hamster}, \texttt{jellyfish}, \texttt{lion}, and \texttt{microphone}.
Finally, for each of the ten classes, we prepare 20 different random latent vectors $z$, with which the generated image $G(\bm{z};\bm{e}_\text{text})$ (without SD-NAE optimization yet) is correctly classified by the ResNet-50.
This step ensures that the initialized sample is not already an NAE, allowing us to isolate the effect and confidently attribute the NAEs to our SD-NAE optimization rather than to other factors inherent in the generative model.

After the preprocessing, we perform 20 * 10 = 200 optimization processes, each corresponding to one class and one prepared latent.
In each multi-step optimization process, if any one of the sample $\bm{x}$ generated at a certain step satisfies $O(\bm{x})=\text{current desired class}\neq F(\bm{x})$, we count this optimization as success in fooling the classifier.
The final fooling rate is calculated as the ratio of successful deceptions to the total number of optimizations, amounting to 200 in our study.
It is noteworthy that we adopt a stricter definition of NAE than that in \cref{eq:nae} to explicitly demonstrate the efficacy of SD-NAE.
In practice, one does not need to enforce $O(\bm{x})=\text{current desired class}$: Even if $O(\bm{x})$ deviates from the expected class, \eg, the synthesized image is not a \texttt{broccoli} while the current text prompt is ``An image of a broccoli'', $\bm{x}$ is still a valid NAE as long as $O(\bm{x})\neq F(\bm{x})$.
Therefore, the fooling rate reported in our experiment represents only a lower bound of the actual fooling rate.

\begin{figure*}[h!]
\centering
\includegraphics[width=0.95\linewidth]{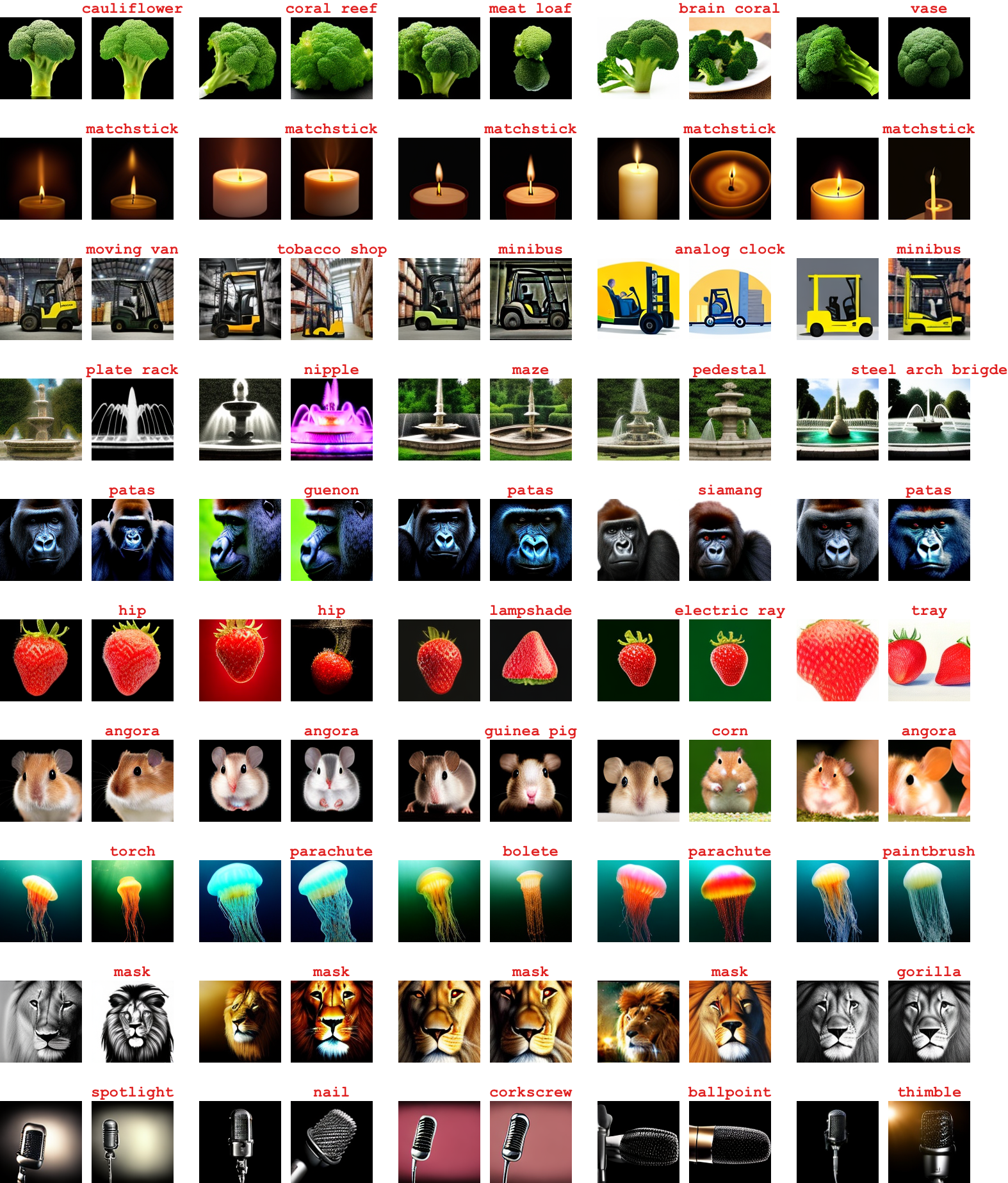}
   %\vspace{-1mm}
   \caption{Examples generated by SD-NAE. From top to bottom, the ground-truth is \texttt{broccoli}, \texttt{candle}, \texttt{forklift}, \texttt{fountain}, \texttt{gorilla}, \texttt{strawberry}, \texttt{hamster}, \texttt{jellyfish}, \texttt{lion}, and \texttt{microphone}, respectively. In each pair, the left one is generated with the initialized token embedding. Importantly, we make sure that all left images are correctly classified by the ImageNet ResNet-50 model in the first place. The right ones are the result of SD-NAE optimization when using the corresponding left one as initialization, and we mark the classifier's prediction in red above the image.}
   %\vspace{-4mm}
\label{fig:results}
\end{figure*}

\textbf{Result.}
As discussed in the main text, SD-NAE achieves a non-trivial 43.5\% fooling rate/success rate. 
The generated NAEs are visualized in \cref{fig:results}.

\section{Comparison with Prior Work}
\label{sec:compare}
Here, we compare our SD-NAE with the method proposed by \citet{uae}.
As mentioned in \cref{sec:related_work}, they perturb the latent vector of class-conditional GANs to curate NAEs, while our design is to optimize the conditional token embedding of Stable Diffusion models (and keep the latent fixed).
In \cref{sec:expr_expand}, we have shown by empirical results that directly applying the previous method (\ie, updating the latent of Stable Diffusion) yields a much worse attack success rate/fooling rate, indirectly justifying our design.
Here, we perform a straight comparison between SD-NAE and the work of \citet{uae}.

\textbf{Setup.}
We use a class-conditional BigGAN \cite{brock2018large} pre-trained on ImageNet, which to our knowledge is one of the most powerful GANs for ImageNet.
The experiment workflow remains the same as with our method: Denoting the GAN as $G$ and the target ResNet-50 classifier we want to attack as $F$, for each image category we prepare 20 randomly-initialized latent vectors $\bm{z}$ where the prediction on each generated image $F(G(\bm{z}))$ matches the ground-truth or the oracle prediction $O(G(\bm{z}))$.
Then using the same loss function as in \citet{uae}, we optimize the latent vector of the GAN to generate NAEs.
We try our best to vary the hyperparameters and report the best result that we observe. 
It is also worth noting that in the original work of \citet{uae}, they only did experiments on small-scale and simple datasets like MNIST, SVHN, and CelebA, while here we are looking at ImageNet with images consisting complex, real-world objects/scenes.

%instead of using the auxiliary classifier loss due to different GAN settings. 
%Targeting the same ResNet50 classifier used in our experiment and assuming a learning rate of 0.1 (the best compared to 0.001, 0.01, 0.05, and 1), we optimize the latent vector of the same ten classes in our method, each with 20 samples that are initially correctly classified, for 20 steps. Following the definition in our paper, we end up with a fooling rate of 17\%, which is much lower than the fooling rate in our method (43.5\%).

\begin{figure*}[h!]
\centering
\includegraphics[width=0.95\linewidth]{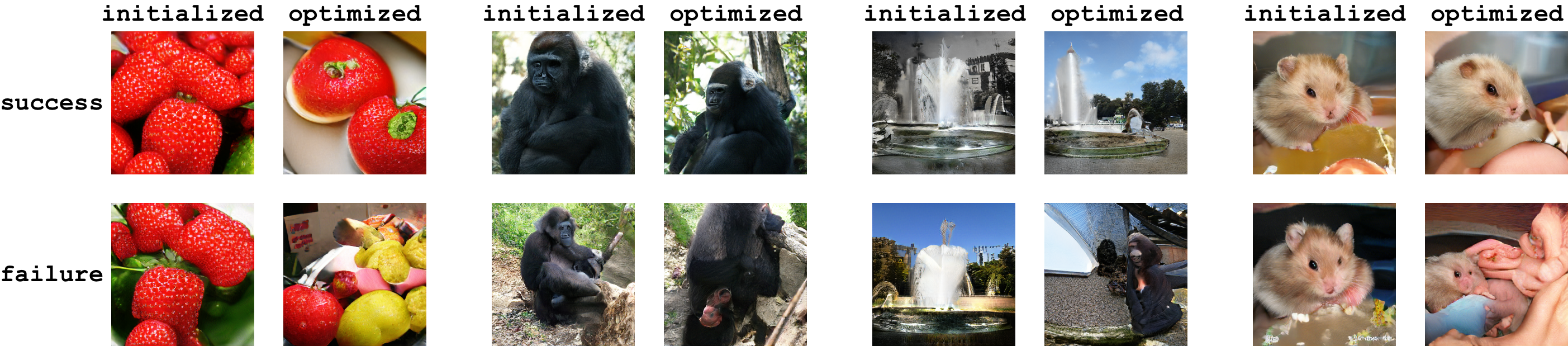}
   %\vspace{-1mm}
   \caption{Examples generated by \citet{uae} using GAN. Following our experiment setup, each initialized image is correctly classified by the target ResNet-50 classifier. The first row shows examples that we count as successfully generated NAEs, whereas the second row shows failure cases where the optimized images exhibit unnatural looking. Note that some successful NAEs here actually do not look that natural, and the quality in general lags behind those generated by SD-NAE (\cref{fig:results}). Still, despite counting them as success, we observe a mere 14.0\% success rate compared with 43.5\% achieved by SD-NAE.}
   %\vspace{-4mm}
\label{fig:gan_visual}
\end{figure*}

\textbf{Result.}
The best fooling rate/attack success rate of \citet{uae} is 14.0\%, which is much lower than ours (43.5\%).
Specifically, in some cases the optimized image does not change much from the initialization and thus fails to deceive the classifier.
In other cases, the optimization goes wild and leads to nonsensical images.
The latter case is in line with the finding that GANs can be sensitive to perturbed latents \cite{dai2023advdiff}, since they might be ``out-of-distribution'' w.r.t. the well-regularized latent distribution that the model sees during the training.
We visualize the synthesized samples in \cref{fig:gan_visual}.
Qualitatively, SD-NAE results in higher-quality samples than the compared method.

%Observing most of the samples generated with \citet{uae}'s method, we find a lower success rate to fool the classifier, and the successfully attacked samples are less meaningful for data augmentation, because $O(\bm{x})$, the semantic ground truth, mostly can only be attributed to classes not in the original class, but less possible to see in reality.

\section{Limitations}
\label{sec:limitation}
Since SD-NAE is based upon Stable Diffusion, it inherits a few limitations from its underlying framework.
First, the computational cost of SD-NAE could be high and the optimization could be slow.
For instance, generating a single 128x128 natural adversarial example with SD-NAE under our experiment setting (\ie, 20 steps for diffusion sampling and 20 steps for SD-NAE's optimization) requires approximately 22GB of GPU memory and takes about 1 minute.
However, we note that both the memory footprint and time cost can be significantly reduced if sampling-efficient diffusion models are used, \eg, Latent Consistency Models \cite{luo2023lcm} and SD-turbo \cite{sauer2023sdturbo} which only require 1 to 4 diffusion sampling steps.
Meanwhile, empirically we find that SD-NAE does \textit{not} really require as many as 20 optimization steps to succeed: In our experiment, the average number of steps for finding the first adversarial example is around 10 (9.66).

%; both the memory footprint and time cost escalate with increased image size and sampling steps.
Second, in some cases, we find that the generated image is absurd and diverges significantly from a natural appearance.
Such cases can arise either inherently from Stable Diffusion or from our SD-NAE optimization process.
Taking the category \texttt{broccoli} as an example, out of the 100 initialization images (generated by Stable Diffusion with random latents), there are 8 of them which exhibits a weird, unnatural looking of a broccoli (an 8\% ``failure'' rate).
Then, during SD-NAE optimization, there are 24 out of 400 images %(20 optimization processes each corresponding to one selected initialization image, and 20 optimization steps within each process, each producing one image) 
that fail to present the normal looking of a broccoli (an 6\% ``failure'' rate).
However, we remark that having unnatural images at a few steps does not mean that SD-NAE is compromised; instead, it can be considered success as long as there is at least one natural-looking adversarial example produced during the multi-step optimization, which is typically the case in our experiment.

%% file: main.bbl
\begin{thebibliography}{15}
\providecommand{\natexlab}[1]{#1}
\providecommand{\url}[1]{\texttt{#1}}
\expandafter\ifx\csname urlstyle\endcsname\relax
  \providecommand{\doi}[1]{doi: #1}\else
  \providecommand{\doi}{doi: \begingroup \urlstyle{rm}\Url}\fi

\bibitem[Brock et~al.(2019)Brock, Donahue, and Simonyan]{brock2018large}
Andrew Brock, Jeff Donahue, and Karen Simonyan.
\newblock Large scale {GAN} training for high fidelity natural image synthesis.
\newblock In \emph{International Conference on Learning Representations}, 2019.
\newblock URL \url{https://openreview.net/forum?id=B1xsqj09Fm}.

\bibitem[Dai et~al.(2023)Dai, Liang, and Xiao]{dai2023advdiff}
Xuelong Dai, Kaisheng Liang, and Bin Xiao.
\newblock Advdiff: Generating unrestricted adversarial examples using diffusion models.
\newblock \emph{arXiv preprint arXiv:2307.12499}, 2023.

\bibitem[Deng et~al.(2009)Deng, Dong, Socher, Li, Li, and Fei-Fei]{deng2009imagenet}
Jia Deng, Wei Dong, Richard Socher, Li-Jia Li, Kai Li, and Li~Fei-Fei.
\newblock Imagenet: A large-scale hierarchical image database.
\newblock In \emph{IEEE Conference on Computer Vision and Pattern Recognition}, pp.\  248--255. IEEE, 2009.

\bibitem[Dhariwal \& Nichol(2021)Dhariwal and Nichol]{dhariwal2021diffusion}
Prafulla Dhariwal and Alexander Nichol.
\newblock Diffusion models beat gans on image synthesis.
\newblock \emph{Advances in neural information processing systems}, 34:\penalty0 8780--8794, 2021.

\bibitem[Geirhos et~al.(2019)Geirhos, Rubisch, Michaelis, Bethge, Wichmann, and Brendel]{geirhos2018imagenettrained}
Robert Geirhos, Patricia Rubisch, Claudio Michaelis, Matthias Bethge, Felix~A. Wichmann, and Wieland Brendel.
\newblock Imagenet-trained {CNN}s are biased towards texture; increasing shape bias improves accuracy and robustness.
\newblock In \emph{International Conference on Learning Representations}, 2019.
\newblock URL \url{https://openreview.net/forum?id=Bygh9j09KX}.

\bibitem[Hendrycks et~al.(2021)Hendrycks, Zhao, Basart, Steinhardt, and Song]{nae}
Dan Hendrycks, Kevin Zhao, Steven Basart, Jacob Steinhardt, and Dawn Song.
\newblock Natural adversarial examples.
\newblock In \emph{Proceedings of the IEEE/CVF Conference on Computer Vision and Pattern Recognition (CVPR)}, pp.\  15262--15271, June 2021.

\bibitem[Ho \& Salimans(2021)Ho and Salimans]{ho2021classifierfree}
Jonathan Ho and Tim Salimans.
\newblock Classifier-free diffusion guidance.
\newblock In \emph{NeurIPS 2021 Workshop on Deep Generative Models and Downstream Applications}, 2021.
\newblock URL \url{https://openreview.net/forum?id=qw8AKxfYbI}.

\bibitem[Luo et~al.(2023)Luo, Tan, Huang, Li, and Zhao]{luo2023lcm}
Simian Luo, Yiqin Tan, Longbo Huang, Jian Li, and Hang Zhao.
\newblock Latent consistency models: Synthesizing high-resolution images with few-step inference.
\newblock \emph{arXiv preprint arXiv:2310.04378}, 2023.

\bibitem[Recht et~al.(2019)Recht, Roelofs, Schmidt, and Shankar]{imagenetv2}
Benjamin Recht, Rebecca Roelofs, Ludwig Schmidt, and Vaishaal Shankar.
\newblock Do imagenet classifiers generalize to imagenet?
\newblock In \emph{International conference on machine learning}, pp.\  5389--5400. PMLR, 2019.

\bibitem[Rombach et~al.(2021)Rombach, Blattmann, Lorenz, Esser, and Ommer]{rombach2021highresolution}
Robin Rombach, Andreas Blattmann, Dominik Lorenz, Patrick Esser, and Björn Ommer.
\newblock High-resolution image synthesis with latent diffusion models, 2021.

\bibitem[Sauer et~al.(2023)Sauer, Lorenz, Blattmann, and Rombach]{sauer2023sdturbo}
Axel Sauer, Dominik Lorenz, Andreas Blattmann, and Robin Rombach.
\newblock Adversarial diffusion distillation.
\newblock \emph{arXiv preprint arXiv:2311.17042}, 2023.

\bibitem[Song et~al.(2018)Song, Shu, Kushman, and Ermon]{uae}
Yang Song, Rui Shu, Nate Kushman, and Stefano Ermon.
\newblock Constructing unrestricted adversarial examples with generative models.
\newblock In S.~Bengio, H.~Wallach, H.~Larochelle, K.~Grauman, N.~Cesa-Bianchi, and R.~Garnett (eds.), \emph{Advances in Neural Information Processing Systems}, volume~31. Curran Associates, Inc., 2018.
\newblock URL \url{https://proceedings.neurips.cc/paper_files/paper/2018/file/8cea559c47e4fbdb73b23e0223d04e79-Paper.pdf}.

\bibitem[Szegedy et~al.(2013)Szegedy, Zaremba, Sutskever, Bruna, Erhan, Goodfellow, and Fergus]{szegedy2013intriguing}
Christian Szegedy, Wojciech Zaremba, Ilya Sutskever, Joan Bruna, Dumitru Erhan, Ian Goodfellow, and Rob Fergus.
\newblock Intriguing properties of neural networks.
\newblock \emph{arXiv preprint arXiv:1312.6199}, 2013.

\bibitem[Zhang et~al.(2023{\natexlab{a}})Zhang, Inkawhich, Linderman, Chen, and Li]{Zhang_2023_WACV}
Jingyang Zhang, Nathan Inkawhich, Randolph Linderman, Yiran Chen, and Hai Li.
\newblock Mixture outlier exposure: Towards out-of-distribution detection in fine-grained environments.
\newblock In \emph{Proceedings of the IEEE/CVF Winter Conference on Applications of Computer Vision (WACV)}, pp.\  5531--5540, January 2023{\natexlab{a}}.

\bibitem[Zhang et~al.(2023{\natexlab{b}})Zhang, Yang, Wang, Wang, Lin, Zhang, Sun, Du, Zhou, Zhang, et~al.]{zhang2023openood}
Jingyang Zhang, Jingkang Yang, Pengyun Wang, Haoqi Wang, Yueqian Lin, Haoran Zhang, Yiyou Sun, Xuefeng Du, Kaiyang Zhou, Wayne Zhang, et~al.
\newblock Openood v1. 5: Enhanced benchmark for out-of-distribution detection.
\newblock \emph{arXiv preprint arXiv:2306.09301}, 2023{\natexlab{b}}.

\end{thebibliography}
